# EVFormer: An Egocentric Vision-EMG Bidirectional Attention Model for Bimanual Hand Pose Estimation

JiaCheng Ge, SiYu Zhang, ShengJie Li, XinTong Yang

## Abstract

Continuous bimanual hand pose estimation from an egocentric viewpoint is fundamental to virtual interaction, wearable control, and rehabilitation assistance. Vision provides hand appearance and spatial structure, but self-occlusion, hand-hand contact, and object manipulation can obscure useful cues. Bilateral wrist surface electromyography (sEMG) reflects muscle contraction and supplies complementary short-term motion information. Because the two modalities differ in sampling, temporal scale, and noise, naive fusion does not necessarily improve accuracy. We propose EVFormer, an egocentric vision-EMG bidirectional-attention model that combines the current RGB image with the preceding 200 ms of bilateral wrist EMG to regress continuous bimanual joint angles. Convolutional and Transformer encoders form visual spatial and EMG temporal representations. Residually scaled sequential bidirectional cross-attention first lets visual features query EMG and then lets EMG features query the updated visual representation; a learnable feature-wise gate fuses the two streams. A shared head predicts 44 finger and wrist angles across both hands. We conduct a single-participant feasibility experiment using one public synchronized EgoEMG recording, partitioned chronologically so that EMG input windows do not overlap across splits. We compare EVFormer with vision-only, EMG-only, late-fusion, and training-mean-pose predictors. Aggregated over 296 test samples and 44 angles, EVFormer achieves a test MAE of 11.482 degrees, outperforming vision-only (13.228 degrees), EMG-only (13.610 degrees), late fusion (13.387 degrees), and the training-mean pose (13.320 degrees). Relative to vision-only and late fusion, the error falls by 13.20% and 14.23%, respectively. EVFormer yields the lowest error in four of the five test gesture classes, but underperforms all three learned baselines for class 21, indicating class-dependent gains. Seed-wise, regional, temporal, and perturbation results that are inconsistent with the corrected aggregate are excluded from quantitative conclusions pending verification against the corresponding raw predictions. These findings provide preliminary evidence that fusing visual spatial cues with EMG temporal information during representation formation can improve pose estimation. Generalization across users, robustness under real occlusion, and online performance remain to be established.

## 1. Introduction

EVFormer studies a specific question: When the first-person camera captures both hands and both wrist electrodes record muscle activity at the same time, can the two signals interact directly in the process of forming posture representation, so as to more accurately predict continuous joint angles? The output of this article is not the gesture category, but 22 angles for the left and right hands. Deviations at every angle are accounted for in the error, so the model must simultaneously handle visible hand shape, short-term muscle activity, and situations where the two are not entirely consistent.

EVFormer is oriented to the first-person vision-myoelectric bimanual posture estimation task and adopts a cross-modal information interaction structure. The current RGB image forms 16 spatial tokens, and the previous 200 ms of wrist electromyography forms 25 time tokens; vision first queries the electromyography, and then the electromyography queries the updated vision, and finally outputs a 44-dimensional angle through feature gating and regression head. This design enables a two-way interaction between spatial cues in the image and recent motion information in EMG before regression.

This design addresses the practical gap of first-person hand observation. The camera can directly record visible finger and wrist shapes, but self-occlusion, hand contact and object manipulation will reduce the available image clues; wrist surface electromyography reflects activities related to forearm muscle contraction, providing another type of motion observation. The sampling methods, time scales and noise sources of the two signals are different. Adding additional EMG will not automatically improve the accuracy. Therefore, the key test of EVFormer is: under the same data partitioning and evaluation indicators, what is the result of the two-way interaction of the feature layer compared to single modality and simple fusion. Figure 1 summarizes the input and hand angle output of this article.

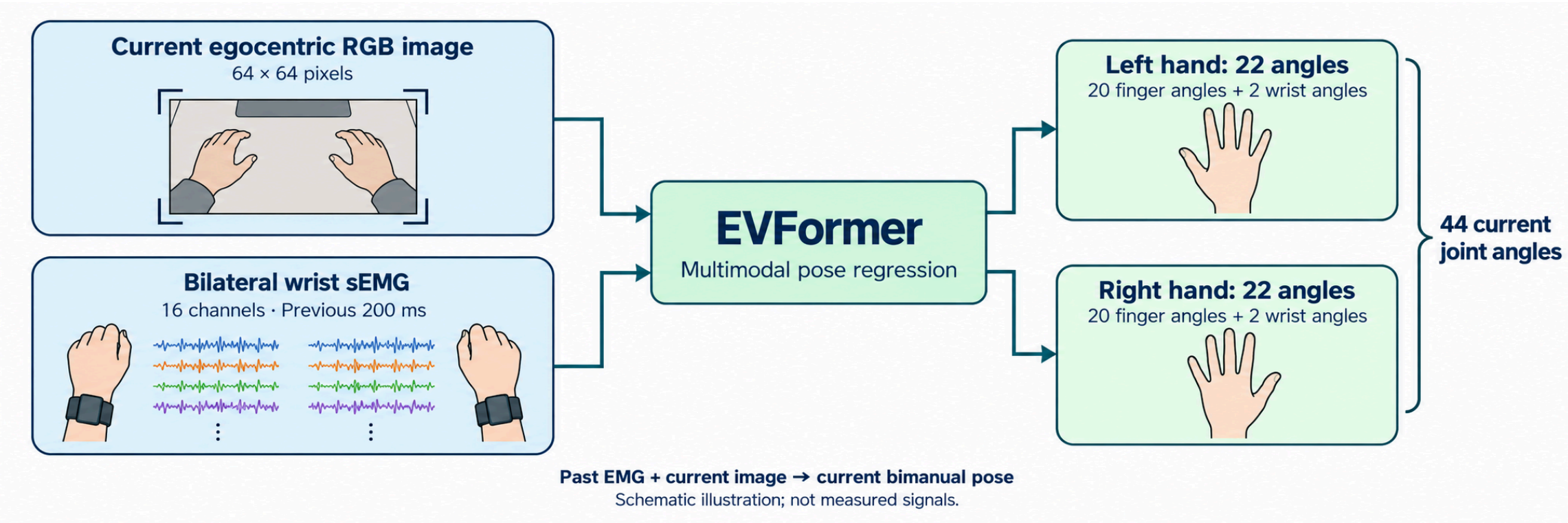

Figure 1. Vision-EMG bimanual hand pose estimation task (schematic). The current image and the preceding 200 ms of bilateral wrist EMG are jointly used to regress the current 44-dimensional joint-angle vector.

This article compares four types of learning models and training mean attitude reference on a public synchronization record. The overall test MAE is 13.228° for vision only, 13.610° for EMG only, 13.387° for late fusion and 11.482° for EVFormer, and the training mean pose is 13.320°. EVFormer has the lowest overall error among the current controls, with an absolute visual reduction of 1.746° and a relative improvement of 13.20%. The principal value is obtained by weighting the five types of gesture results in Table 6 according to the number of samples. Category analysis retains the counterexamples of Category 21; the original sub-, region-, and time-based grouping and perturbation results are inconsistent with the corrected main values and need to be re-aggregated. It is not claimed that each group or each random seed has improved based on this.

The task background comes from EgoEMG [1]: it provides synchronized first-person video, bi-wrist electromyography and continuous joint angles of both hands, and establishes evaluation tasks from electromyography to posture, vision to posture and two-modal fusion. EMG2Pose [2] previously provided the basis for large-scale research on bilateral wrist electromyography and hand movements. This paper uses EgoEMG's target semantics to study EVFormer, but the currently available experiments only involve a public mirror recording of one of the participants, so it cannot replace the original work with new users or a complete multi-partition evaluation. We explain the synchronization, filtering and partitioning methods of this record in the methods and experiments section.

The main contributions of this article include three aspects. First, a 44-dimensional angle regression model for both hands is constructed for synchronized first-person images and bi-wrist electromyography, using residual scaling sequential bidirectional cross-attention and feature-by-feature gating to exchange spatial and temporal information before regression. Second, comparing single-modality, late-stage fusion and complete models on fixed single-segment divisions, uniformly summarized by sample-weighted MAE, EVFormer achieved the lowest overall error of 11.482° in the current control. Third, the differences between overall gains and local failures are preserved through gesture category analysis, and the sub-results that need to be reviewed and the conditions required for a complete multi-user evaluation are clarified.

## 2. Related Work

### 2.1 Surface EMG and Continuous Hand Pose Estimation

Surface electromyography records the electrical activity of muscles on the surface of the skin. Farina et al. discuss the physiological basis for analyzing neuromuscular control signals from surface electromyography [15]; this observation has a different physical origin than the visible hand shape recorded by a camera. Early continuous regression studies have estimated multiple finger joint angles from muscle activation, not just discrete gesture categories [12,13]. NeuroPose uses wearable electromyography for three-dimensional hand posture tracking [14], and EMG2Pose provides larger-scale bi-wrist electromyography and continuous joint angle benchmarks [2]. These works illustrate that EMG to posture can be studied, but electrode placement, subject differences, and task changes can still change signal distribution.

The goals and size of the data set determine the questions that can be answered. Ninapro [11] and GrabMyo [19] mainly support electromyographic gesture recognition, and EMG2QWERTY [20] is oriented to typing tasks with wrist electromyographic input; Andreas et al. [16] and Jarque-Bou et al. [17] jointly record electromyography and hand movements in daily activities. EgoEMG [1] further provides synchronized first-person images on a continuous bimanual joint angle task. EMGBench [18] places special emphasis on cross-distribution generalization and adaptation: Consecutive time-block results for the same person cannot substitute for new user evaluation. The public record used in this article is suitable for verifying synchronization and model behavior, but cannot measure cross-user changes.

In terms of modeling, VQ-MyoPose explores vector quantized myoelectric pose representation [21], emg2tendon connects myoelectric decoding to tendon drive control [22], and DexEMG focuses on myoelectric to pose generalization [23]. Recent myoelectric representation learning also includes scalable Transformer [42] and lightweight base models for edge devices [43]. These methods are not identical to the dual-modality setting of this article, but suggest that EMG encoding capabilities, training scale, and cross-user stability need to be examined separately. EVFormer's current 200 ms input window and one-layer timing encoder are lightweight implementations; this article does not rank it directly with the heterogeneous data set results of the above methods.

## 2.2 Egocentric Vision, Bimanual Interaction, and Angle Representation

MANO provides parametric models for hand shape and posture [7]. FreiHAND [24] supports single-frame RGB hand reconstruction, InterHand2.6M [25] focuses on interacting hands, and HO-3D [26] and ARCTIC [27] further incorporate handle and object interaction into three-dimensional perception. The first-person environment also includes data sets such as H2O [28], DexYCB [29], HOT3D [9]; their camera positions, annotation forms, and evaluation goals are different, and angle MAE cannot be directly compared just because they all contain hand images. UmeTrack [8] demonstrates a multi-view tracking solution for near-eye devices.

Hand reconstruction models can use either specialized geometry and detection pipelines or a general-purpose vision backbone. HaMeR [30] and WiLoR [10] target hand positioning and reconstruction; GigaHand [31] expands the annotation scale of first-person and external perspectives. ResNet [32], Visual Transformer [4] and DINOv2 [33] represent different general feature extraction routes. EVFormer's image branch learns a 64×64 fixed region from random weights, and its results can only be interpreted as internal comparisons under the same preprocessing; pretrained vision models may significantly change the single-modal baseline and fusion space. The 44-dimensional target of this article is the joint angle of both hands, and the results cannot be directly converted with the method that only outputs three-dimensional key points or MANO parameters.

## 2.3 Where to Fuse Vision and EMG

Co-modeling of vision and electromyography has been used for gesture understanding [38] and grasp intention inference [37]; cross-modal pre-training has also begun to link electromyography and posture information [36,41]. Residual fusion of EgoEMG uses electromyography as a correction for visual angle prediction [1]. EVFormer advances the fusion to the feature layer: tokens are formed according to visual positions and electromyographic time slices, information is exchanged bidirectionally through scaled dot product attention [3], and then angles are returned through gating. The design of feature gating can be linked to the idea of channel recalibration [35], but the gating in this paper has no visibility or physiological confidence supervision and thus cannot be interpreted as a calibrated sensor quality score.

Different fusion locations do not necessarily mean fusion benefits. EgoEMG has reported residual improvements in matching the visual backbone [1]; the average test error of EVFormer in the current single segment of this article is 11.482°, which is lower than the 13.228° in the visual branch. The two observations come from different data sizes and training conditions. The appropriate comparison objects should be vision, electromyography, late fusion, residual fusion and two-way interaction models under the same division, and the effective parameter amounts should be recorded at the same time. This article uses AdamW [34] to train a lightweight model. The current overall advantage belongs to the internal control results, and the mechanism contribution needs to pass parameter matching and ablation testing.

## 2.4 Motivation for EVFormer

EVFormer's design revolves around the complementarity of different modal information: current images provide spatial cues, and historical EMG provides recent movement cues. The model adopts the sequential bidirectional cross attention of vision → electromyography, electromyography → updated vision, and controls the contribution of the two types of representations to the final prediction through feature gating. The output is a first-person image and a historical EMG joint regression of 44 current angles of the hands.

In the current single-segment experiment of this article, the overall test MAE of EVFormer is 11.482°, which is 13.20% lower than the visual only, and has the best overall performance among existing controls; this result still needs to be reviewed on the official multi-user division.

### 2.5 Relationship to EgoEMG and Research Questions

This paper follows EgoEMG's simultaneous visual-myoelectric input and continuous hand angle targets to study whether advancing the fusion to the posture representation formation stage can improve the overall error of the current single segment. EVFormer uses the same test center and 44-dimensional target as the internal baseline of this article; the modified sample-weighted MAE supports its overall advantage, but cannot replace the evaluation of the EgoEMG residual fusion baseline or official multi-user partitioning, nor can it prove improvement in every category and every seed.

## 3. Dataset and Pose Annotations

### 3.1 Data Source and Bimanual Angle Targets

The EgoEMG paper describes the simultaneous acquisition of bilateral eight-channel wristband, first-person RGB, external video, inertial sensor and optical motion capture data [1]. The targets for each hand are 20 finger angles arranged in UmeTrack/EMG2Pose order, as well as two angles of wrist flexion and extension, and radioulnar deviation. EVFormer predicts both hands jointly, so the total output is 44 scalars: positions 0-21 belong to the left hand, and 22-43 belong to the right hand. The main metric is the absolute angular error averaged over the valid test sample and output dimensions, in degrees. The data scale cited in this article and the benchmark values of the original paper correspond to arXiv version 1; the continuously updated repository may release subsequent data versions.

The pre-experiment used the third-party public image IInfact/EgoEMG-snippet [5], which only contains a segment of EgoEMG acquisition. The Parquet file contains 691,374 lines, covering approximately 345.7 s; the corresponding camera video has 20,740 frames, the frame rate is 60 fps, and the screen size is 1280×720. The subject_id and episode_index fields only have values 29 and 28 respectively. Single segments cannot estimate uncertainty across participants.

### 3.2 Synchronization, Preprocessing, and Label Filtering

Candidate prediction centers are selected every 200 rows; for a nominal 2 kHz myocurrent, this corresponds to approximately 10 candidate samples per second. The model uses 400 consecutive EMG sampling points before the center row, excluding the center point itself. The left and right filtered EMGs each have 8 channels and are spliced into 16 channels; every 4 consecutive original time sampling points are averaged to form a time series with a length of 100. The current image is selected based on the camera frame index in the center row. After decoding the corresponding video frame, a fixed area in the lower center of the screen is captured and scaled to a 64×64 RGB image. This region was determined for this pilot experiment after examining the video footage and is not the hand crop used in the EgoEMG official visual baseline.

The finger angle array in the published file is in radians. This article converts the 20 finger angles of each hand into degrees; the wrist pitch and yaw fields have been stored in degrees and will not be converted when splicing. Candidate samples will be retained only when the labels generated by the left and right hands are valid, the left and right wrist angle flags are valid, the image frame index is non-negative, and the EMG and target values are both limited. After checking the labels and inputs, 1,968 synchronized candidate samples were obtained; after dividing the boundaries and removing 4 samples that may cause window overlap, the final number of samples used in the model was 1,964. The mean and standard deviation used for EMG normalization were calculated from training blocks only and used as-is for validation and testing. Figure 2 shows the preprocessing process and input time relationship of synchronized samples.

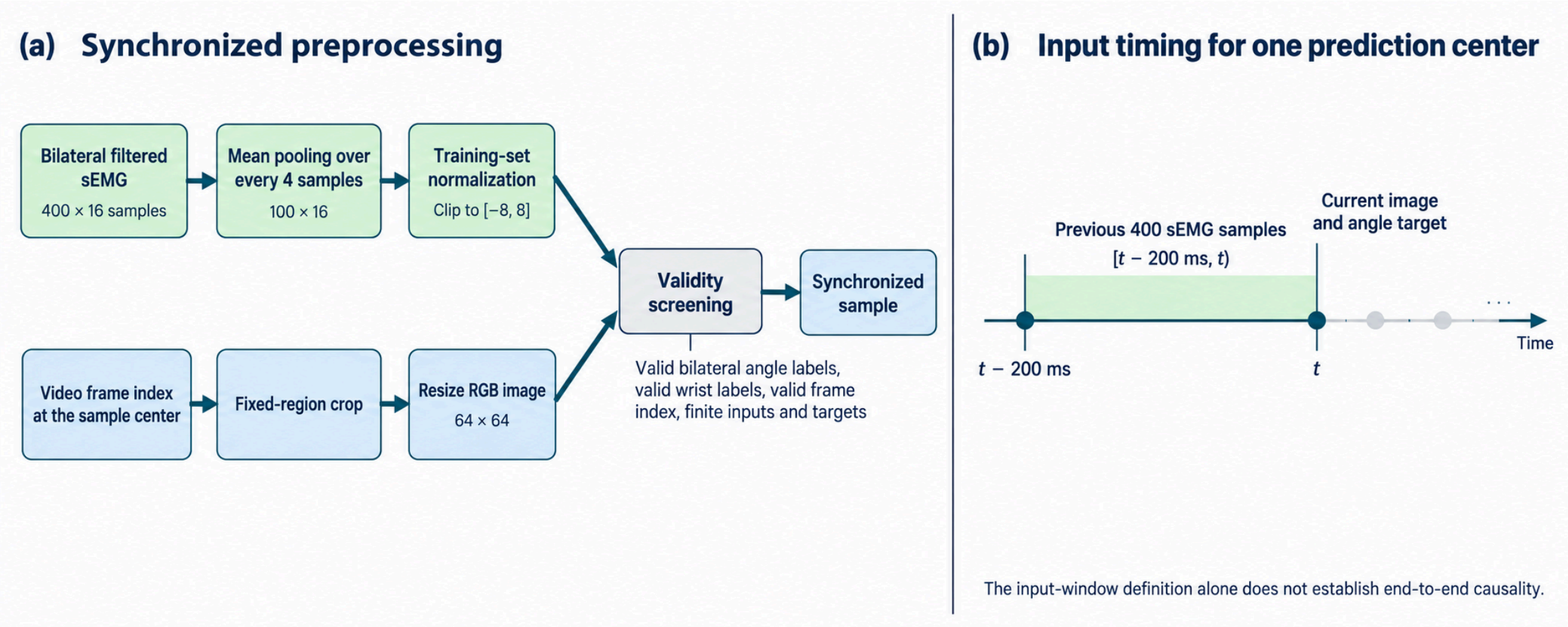

Figure 2. Synchronized sample construction and preprocessing (schematic). (a) EMG binning, normalization, and image cropping. (b) A 200 ms window preceding the prediction center and excluding the center sample. The input-window definition alone does not establish end-to-end causality.

### 3.3 Dataset Size and Experimental Splits

The valid samples are sorted by the center row index, and the order is approximately divided into 70%/15%/15%. Each boundary removed enough samples that the 400-point EMG window overlapped with adjacent data blocks, resulting in 1,375 for training, 293 for validation, and 296 for testing. The training time is 0.2-169.7 s, the validation time is 170.0-199.7 s, and the testing time is 200.0-281.0 s. Input EMG windows for different data blocks do not overlap. This avoids the most direct window overlap leaks, but does not allow data chunks to become independent of each other or acquisition sessions.

The distribution of gesture categories differed significantly between time blocks. A total of 1,375 valid gesture samples were trained, 47 of which did not have gesture category labels, and the remaining labeled categories were 8-19; 19, 20, and 21 for verification; and 21, 22, 23, 24, and 27 for testing. None of the test categories appear in the training block, but category 21 has appeared in the validation block for checkpoint selection. Therefore, this test reflects changes in temporal and category distribution under the same participant, and all test categories cannot be collectively referred to as gestures not seen throughout the model selection process. It is also not an official EgoEMG cross-gesture breakdown, and both time and category changes need to be taken into account when explaining the gap from verification to testing. Figure 3 shows both the sample order division and the physical time range.

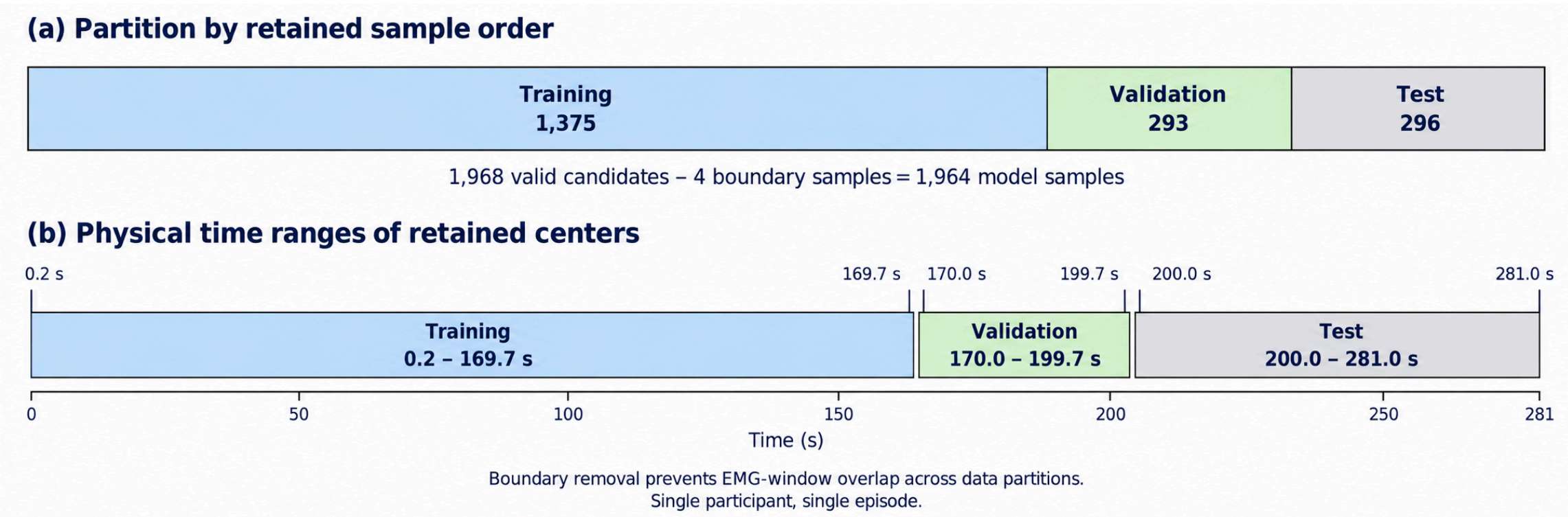

Figure 3. Single-recording data split (schematic). (a) Sequential split of retained sample centers and sample counts. (b) Physical time ranges of the retained centers in each split. The upper and lower rows show sample counts and physical time, respectively.

### 3.4 Data Quality and Scope

This public clip image is not an official multi-participant benchmark resource. The video screen contains the background of the acquisition device, and the fixed cropping area is small; many lines later did not pass the tag validity filter, so although the source file lasts until about 345.7 s, the last valid test sample in this article is only located at 281.0 s. Published filtered EMG may also be processed offline. Merely using the window before the center point does not prove that the entire processing chain meets the requirements for real-time, causal deployment. To evaluate online systems, causal filtering needs to be explicit and account for timestamps, video decoding, and inference delays.

# 4. Unified Tasks and Model Baselines

## 4.1 Three Unified Tasks

Referring to the task definition of EgoEMG [1], this paper uses the same batch of synchronized samples and the same 44-dimensional angle target for three conditions: from electromyography to posture, only observing the signals of both wrists, from vision to posture, only observing the current RGB image, and from electromyography to visual fusion, observing two inputs at the same time. The target moment, angle unit, and valid label mask of the three conditions were consistent with the test center. This way, the difference in results at least does not come from different test frames or different output dimensions. The 22-dimensional target for each hand in the original EgoEMG article semantically corresponds to the joint 44-dimensional output of both hands in this article, but the training examples and data partitioning are not the same.

Let the central time be t, the image be I_t, the EMG of both wrists at the first 400 sampling points be E_[t-400:t), and the supervision target be y_t∈R^44. The three prediction functions are written as f_e(E), f_v(I) and f_ev(I,E) respectively. All reported errors are obtained by first restoring the network output to angles and then averaging the absolute errors over the 44 corners of the valid sample. This task is continuous pose regression; it cannot demonstrate discrete "movement intent" recognition performance based on angular error alone.

## 4.2 Evaluation Protocol and Training Settings

The main indicator takes the average absolute error at all valid test centers and 44 angles of each random seed, and then averages it among seeds: MAE=(1/S)∑s[(1/(N×44))∑i∑d|ŷs,i,d−yi,d|]. When the category or time grouping covers the same test set and uses the same output dimension, its overall MAE must be weighted by the number of samples in each group; when the left and right hands have 22 dimensions each, half of it, and the 40 dimensions of the fingers and the 4 dimensions of the wrist are weighted by 40:4. This article uses the sample weighted result 11.482° in Table 6 as the EVFormer main value. The relative improvement rate is (MAE baseline − MAE_EVFormer)/MAE baseline × 100%. The difference and percentage in the table are calculated based on the displayed three-digit decimal main value. The sub-results and test standard deviations of EVFormer that cannot match the main value are not reported for the time being.

The target angle is divided by 90 during training and a Smooth L1 loss of beta=0.05 is used in normalized units. The optimizer is AdamW, learning rate $1.5\times10^{-4}$, weight decay $5\times10^{-4}$, batch size 32, gradient norm clipping threshold 1.0. The vision branch starts with random weights and does not load ImageNet or hand pose pre-training checkpoints. Training is performed for a maximum of 60 epochs; the first 5 epochs are subject to learning rate warm-up, and then cosine decay is used; if the verification MAE does not improve for 10 consecutive epochs, it will stop early. Training uses a Transformer dropout of 0.10, a modality dropout of 0.10, a bidirectional cross-attention residual scaling coefficient of 0.10, and the gated last layer bias is initialized to +2.0. For each condition and random seed, the checkpoint with the lowest verification MAE is selected and evaluated once on the test set. The three seeds of 2026, 2027 and 2028 control the model initialization and data reading order, while the data partitioning remains the same.

## 4.3 EMG-to-Pose Branch

The 100×16 binned EMG sequence is first transposed and then input into a one-dimensional convolution. Two convolutional layers with GELU activation and stride 2 shorten the sequence to 25 64-dimensional tokens. Learnable temporal positional embedding of token ordering, a layer of four-head Transformer encoders integrates information over short histories. All EMG channels were normalized using the training block statistic and the normalized absolute values were truncated to within 8. This pre-experiment directly uses the filter field provided by the data set and does not learn the original signal filter in the network.

## 4.4 Vision-to-Pose Branch

The 64×64 RGB image is sequentially passed through two convolutional layers with stride 2 and GELU activation. Adaptive pooling generates 4×4 feature maps, which flatten the spatial dimensions to form 16 64-dimensional visual tokens. Each token is added with a learnable position embedding. Subsequently, a layer of Transformer encoder containing four attention heads, 128-dimensional feedforward sub-layer performs intra-image self-attention. This lightweight visual backbone is designed for single-segment feasibility experiments and cannot be considered a competitive alternative to the larger pre-trained visual backbones in EgoEMG [1] or WiLoR [10].

### 4.5 Vision-EMG Fusion: EVFormer

EVFormer includes four stages: dual-channel encoding, sequential two-way interaction, feature gated fusion and 44-dimensional angle regression, and uniformly uses 64-dimensional features. The visuospatial token and the electromyographic time token undergo bidirectional cross attention in sequence, and then form a regression representation through feature gating. The camera cropping includes two hands at the same time, and the electromyography inputs the data of both wrists in a fixed channel sequence. The overall structure is shown in Figure 4.

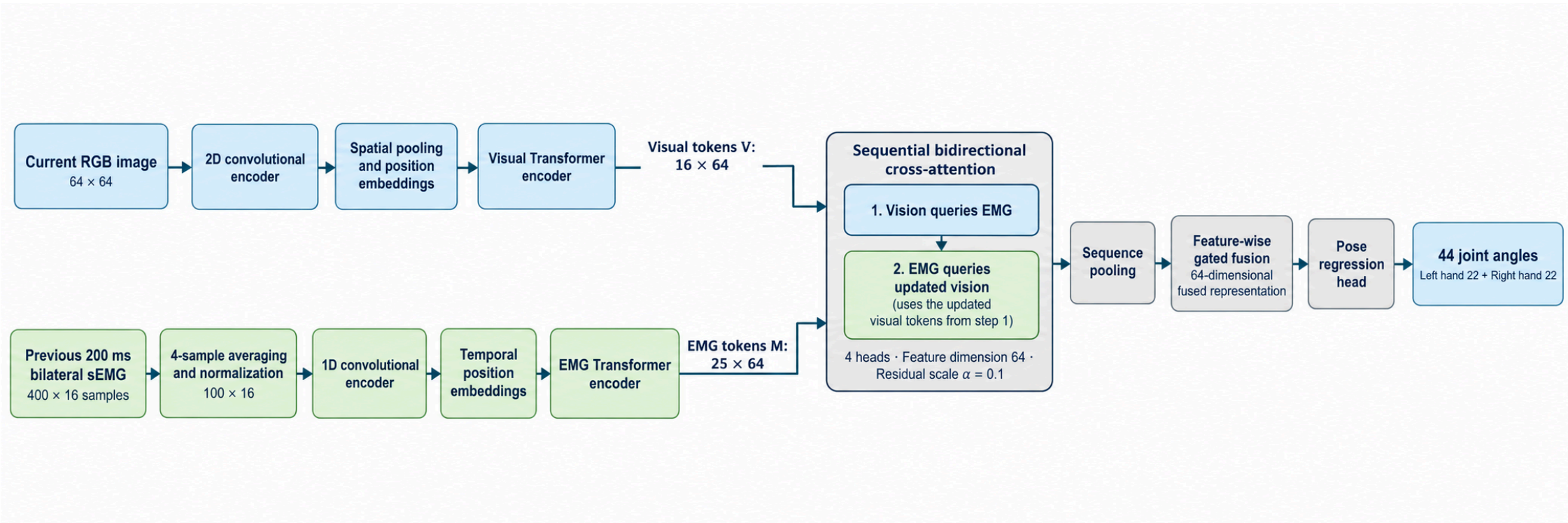

Figure 4. EVFormer architecture with 64 x 64 images, a 200 ms EMG history, residual scaling, bidirectional cross-attention, and gated fusion.

Given a query Q, keys K, and values U, scaled dot product attention is written as $A(Q,K,U)=\mathrm{softmax}(QK^T/\sqrt{d})U$ [3]. EVFormer first lets the vision query the electromyography and uses the residual scaling coefficient $\alpha=0.1$ to get $V'=V+\alpha A(V,M,M)$; then it lets the electromyography query the updated vision and gets $M'=M+\alpha A(M,V',V')$. Both operators use four attention heads. The second interaction uses the updated visual sequence, so it is a sequential two-way interaction rather than two simultaneous, independent interactions. Residual scaling is used to reduce the perturbation of random cross-modal attention on single-modal representation in the early stage of training.

The attention cost is related to the number of two-way tokens. In this preliminary experiment, 16 visual tokens and 25 electromyographic tokens formed 400 pairs of queries and keys in each direction and each attention head; there were also 400 pairs in the reverse direction. This is less distracting than directly computing all 400 EMG raw sample points or pixels of the entire image. However, we cannot directly claim that the system has low deployment latency because video decoding, preprocessing, convolution and sensor sampling also affect the end-to-end time consumption.

The two-layer perceptron generates 64-dimensional gating $g=\mathrm{sigmoid}(\mathrm{MLP}([v,m]))$, and its last layer bias is initialized to +2.0 to bias the initial gating toward visual features; the fusion weights of each feature dimension are then learned from the training data. The fusion vector is $z=g\odot v+(1-g)\odot m$, where $\odot$ represents element-wise multiplication. Layer normalization and a two-layer regression head map z to 44 joint angles. Gating is not supervised by visibility or physiological signal quality labels and therefore cannot be interpreted as a calibrated sensor reliability score. Figure 5 details the sequence relationship and gating calculation of the two interactions.

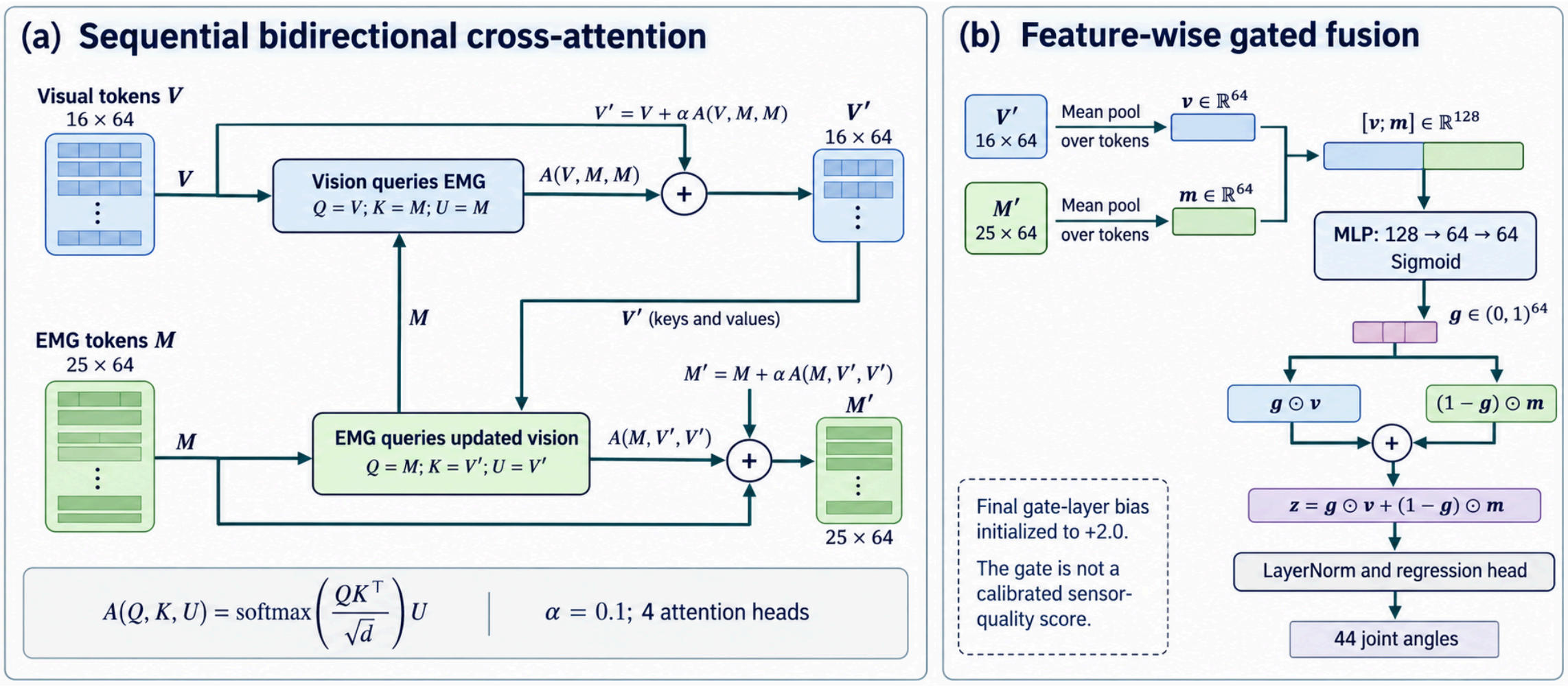

Figure 5. Sequential bidirectional cross-attention and feature-wise gating (schematic). (a) Vision first queries EMG, after which EMG queries the updated visual representation. (b) The two summary vectors are fused by a 64-dimensional gate and mapped to 44 angles. The gate is not a calibrated sensor-quality score.

Only the visual variant feeds the visual summary vector into a common pose head; only the myoelectric variant uses the myoelectric summation vector. The late fusion retains the two-way encoder and gating, but skips the two cross-attention operators; the complete EVFormer adds these two cross-attention operators. The four conditions used the same image preprocessing, EMG windows, targets, optimization settings, and time divisions. Each condition is retrained at each random seed. The code builds the complete module in all modes, so directly reporting the total number of parameters for the module will overestimate the number of parameters actually involved in the calculation in the single-mode condition; this article will not use "the same nominal parameters" as a performance argument.

The fusion baseline of EgoEMG first obtains the visual joint angle, and then uses electromyographic features to predict the angle residual, which belongs to the output layer residual correction [1]. EVFormer exchanges space and time tokens before angle regression, and then uses gated synthesis to represent them. Its structural features compared with the residual path are: the image area and the EMG time slice can be queried each other, the fusion occurs during the formation of the posture representation, and the angles of both hands are jointly returned by the same output head. The post-fusion comparison in this article is used to measure the effect of token interaction; to compare the prediction accuracy of these two fusion mechanisms, it is also necessary to run the residual baseline under the same partitioning, image cropping and training budget.

### 4.6 Connection to Formal Benchmarks

Model training and evaluation are completed by the computing server. The existing materials do not provide the specific hardware, driver and software version of the server. The corresponding configuration and operating environment need to be supplemented when officially archiving. Unconfirmed other equipment environments are not used as experimental platform information in this article.

To conduct a rigorous model comparison, it is necessary to use the EgoEMG official complete data and its three predefined test divisions of "gesture", "user" and "both at the same time" [1]. All methods in the same comparison should adopt consistent single-handed 22-DOF targets, effective frame rules, center frame supervision, hand image cropping, and results aggregation. At a minimum, official visual baselines, EMG baselines, residual fusion baselines, and EVFormer and post-fusion controls of comparable size are required. Pre-training, data augmentation and training duration must be clearly disclosed. The original paper summarizes the results on different partitions and reports the degree of dispersion among users; single-segment time partitioning cannot replace this protocol. Table 1 distinguishes the experiments performed in this paper from the formal comparisons required in the future.

Table 1. Single-fragment pilot and formal evaluation protocol

| **Assessment project** | **Single fragment pre-experiment** | **Required for formal assessment** | **Current status** |
|---|---|---|---|
| participants | Only one person numbered 29 | Official participant division | Pre-experiment only |
| Image input | Fixed 64×64 camera crop | Official hand cut | Settings are different |
| EMG history | 200 ms before | Use consistent windows across models | Pre-experiment only |
| generalization axis | continuous time block | Gesture, user, and both | Not yet completed |
| control model | Vision, electromyography, post-fusion | Published residual baselines are also required | incomplete |
| uncertainty | Three initialization seeds | Cross-user and cross-seed changes | incomplete |

This article uses the 11.482° weighted by the number of samples in Table 6 as the EVFormer overall test MAE, and unifies the main table and text accordingly. The original seed-by-seed, region, time grouping and perturbation values that are inconsistent with this value have been removed and items to be reviewed are marked with "-"; the corresponding standard deviation, winning frame ratio and grouping improvement conclusions are not reported yet. Existing archived material does not contain matching piece-by-piece seed JSON, checkpoints, raw frame-by-frame predictions, or logs, and these piece-by-piece results cannot be independently recovered. A complete review requires obtaining the corresponding products on the experimental server. Sections 7.5-7.6 are the unfinished follow-up experimental plans.

The original file identity record is as follows: Parquet fragment SHA-256 downloaded this time Photographed by 551ee76cb78f11c9f64c211f6f236737b7d74cbac0d9dc18ddba3d0641efb602 The camera video is 88df1ba9c72f690b7277cead6c17ad23c0126a004b4de68114860effbaf916d6. The hash can accurately identify the two source files used in the pre-experiment, but it cannot prove that the third-party image is exactly the same as every released version of the official EgoEMG. Formal benchmarks should use official repositories and data descriptions [6].

# 5. Experiments

## 5.1 EMG-to-Pose

On the same test center, the average MAE of the three seeds of the EMG-only model was 13.610° with a standard deviation of 0.485°. The verification MAE is 10.691°; the increase in test error and seed fluctuation indicate that the current branch is not stable for single-segment back-end data. The EMG model uses a 200 ms history and a lightweight encoder, and its results belong to the internal baseline of this article and cannot be directly ranked with the EMGFormer formal score of the EgoEMG original article. The finger and wrist errors are 13.222° and 17.492° respectively. The wrist has only four output dimensions, and a single total MAE tends to weaken its higher error.

Table 2. Verification and overall test MAE under unified division

| **method** | **Validation MAE (degrees)** | **Test MAE (degrees)** | **Relative vision test difference (degrees)** |
|---|---|---|---|
| training mean pose | Not applicable | 13.320 | +0.092 |
| visual only | 10.415 ± 0.021 | 13.228 ± 0.049 | 0.000 |
| EMG only | 10.691 ± 0.039 | 13.610 ± 0.485 | +0.382 |
| Post-fusion without cross attention | 10.627 ± 0.069 | 13.387 ± 0.419 | +0.159 |
| EVFormer | 8.720 ± 0.080 | 11.482 | −1.746 |

Table 2 summarizes the overall results, with ± followed by the standard deviation of the three training seed samples in the original report. The overall test MAE of EVFormer adopts the sample weighted correction value of 11.482° in Table 6; the test seed standard deviation matching this principal value has yet to be reviewed, so this cell does not have a ± value attached. The validation set scores are temporarily retained as the original reported values and are not used to derive the test standard deviation. EVFormer is lower than vision only 13.228°, electromyography only 13.610°, late fusion 13.387° and training mean attitude 13.320°, making it the overall best model in the current comparison. The training mean posture differs from vision by only 0.092°, and the single-participant results are still insufficient to infer stable effects at the population level.

### 5.2 Vision-to-Pose

The average test MAE of the visual branch alone is 13.228°, and its finger and wrist MAEs are 12.817° and 17.343° respectively. The overall MAE after EVFormer's correction is 11.482°, with a relative visual reduction of 1.746°. The original EVFormer finger and wrist summary is inconsistent with this main value, and its regional level values are not reported yet. Fixed 64×64 cropping, single participant, and single episode limit the scope of the current conclusions.

Table 3 retains the left and right hand, finger and wrist errors of the three baselines. The original area results of EVFormer cannot reconstruct 11.482° according to the left and right hand or 40:4 dimension weighting, so it is marked with "—" to be reviewed by the matching original prediction. This symbol indicates that no consistency-checked values are reported and does not indicate zero error. The wrist only accounts for 4 of the 44 output dimensions. The regional results need to be summarized by dimension, and the fingers and wrist cannot be simply averaged.

Table 3. MAE of left and right hands, fingers and wrists (EVFormer sub-items to be reviewed)

| method | Left hand MAE | Right MAE | Finger/wrist MAE |
|---|---|---|---|
| visual only | 13.264° | 13.192° | 12.817° / 17.343° |
| EMG only | 13.587° | 13.633° | 13.222° / 17.492° |
| late fusion | 13.310° | 13.465° | 12.984° / 17.419° |
| EVFormer | — | — | — / — |

EVFormer's left and right hand, finger, wrist and joint-by-joint results need to be re-aggregated from predictions consistent with the primary values. The current low overall MAE cannot directly infer that each hand, each type of joint, or each degree of freedom has been improved. This article will not make such sub-conclusions for the time being.

### 5.3 Fusion of EMG and Vision

The average test MAE of EVFormer is 11.482°, which is lower than vision only 13.228° and late fusion 13.387°. The absolute relative visual reduction is 1.746°, the relative improvement is 13.20%, and the overall error is the lowest among all current comparisons. This result supports the overall fusion benefits in the current single-segment setting, which still needs to be verified by a full multi-user evaluation.

Post-fusion retains the dual-channel encoder and gating, but does not perform token layer exchange. The average test MAE is 13.387°; EVFormer is 11.482°, with an absolute reduction of 1.905° and a relative improvement of 14.23%. Compared with only the EMG and training mean posture, EVFormer decreased by 2.128° and 1.838° respectively, a relative improvement of 15.64% and 13.80%. The effective training parameters of late fusion and EVFormer are different. This comparison cannot yet attribute all the benefits to cross attention alone, and requires parameter matching and ablation verification. Figure 6 provides a visual comparison of the overall test MAE in Table 2.

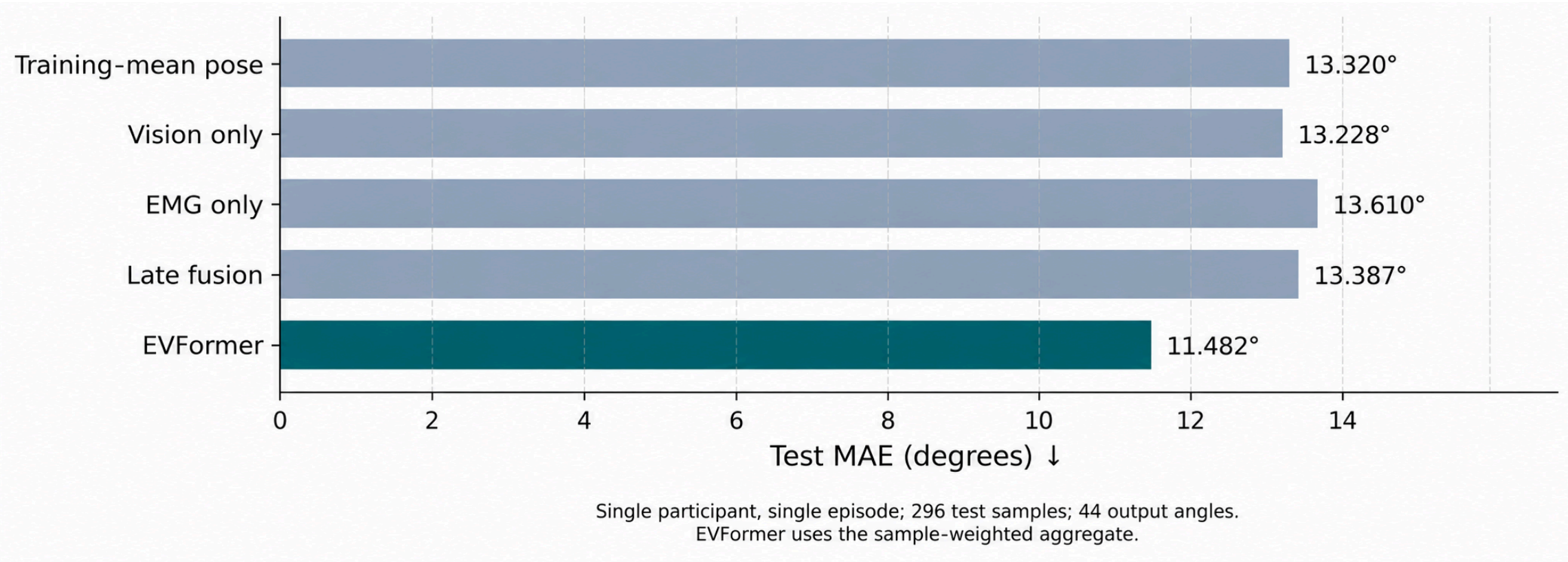

Figure 6. Overall test-MAE comparison. Values are taken from Table 2 using the same 296 test centers and 44-dimensional targets; the EVFormer result is the sample-weighted primary value reconstructed from Table 6. Error bars are omitted pending verification.

The overall MAE describes the average error of all test angles and cannot directly give the benefit frame ratio. The EVFormer seed-by-seed results in the original report are inconsistent with the revised main values, and the original winning frame ratio is no longer used as matching evidence for the current results. During review, the 44-dimensional errors of the same seeds, checkpoints and test centers need to be compared frame by frame.

After the matched frame-by-frame predictions are archived, the pairwise error difference, winning ratio, median, and quantile of the EVFormer and the visual baseline should be calculated. The proportion of frames without review is currently not reported, nor is the overall error reduction interpreted as an improvement in each frame; specific benefit scenarios require independent visibility and signal quality annotation.

### 5.4 Sensitivity to Input Changes

Table 4 lists the controlled input disturbance conditions to be reviewed, and takes 11.482° as the corrected clean input overall MAE reference. The clean value used in the original disturbance report is inconsistent with the corrected main value, and the same checkpoints and test predictions have not been confirmed, so the original disturbance absolute value and relative change have been removed. Except for clean references, "—" indicates that the corresponding results need to be reviewed, and the quantitative impact of local or entire input transformations cannot be derived based on this.

Table 4. EVFormer input disturbance review conditions (only clean overall reference is listed)

| **Test input conditions** | **EVFormer MAE (degree; items to be reviewed are marked with -)** |
|---|---|
| clean input | 11.482° (clean reference) |
| Center 32×32 pixels of image zeroed | — |
| Image 5×5 blurred | — |
| Set an electromyographic channel to zero on each wrist | — |
| Normalized EMG with 0.5 standard deviation Gaussian noise | — |
| The center of the image is set to zero and one channel on each wrist is set to zero. | — |
| Set the whole channel image to zero | — |
| Set the whole circuit of myoelectricity to zero | — |
| The image and EMG are reset to zero at the same time | — |

The perturbation review should fix the 296 test centers and original angle labels, use checkpoints consistent with the clean master results, evaluate central image zeroing, blur, channel zeroing, EMG noise and combined conditions separately under each seed, and then summarize the mean and relative changes. There is currently no perturbation result matching 11.482°. It is not claimed that mild perturbation will only increase the error slightly or that removing the entire path will inevitably increase the error.

Subsequent cross-sectional robustness comparisons require applying the same applicable perturbations to the visual, electromyographic and fusion baselines and exposing checkpoints and predictions. The central fixed rectangle does not directly represent hand-by-hand true occlusion, zeroing out the standardized input does not equate to physical device damage, and software diagnostics are not a substitute for real conditions or cross-user evaluation.

### 5.5 Time Periods and Data Distribution

The validation MAE of vision only, EMG only and late fusion is 10.415—10.691°; EVFormer is 8.720° (three seeds standard deviation 0.080°). The test error of each learning model is higher than the validation error. Gesture categories vary along time blocks: training focuses on categories 8–19, validation on 19–21, and testing on 21–24 and 27. However, the validation-to-test gap cannot be entirely attributed to unseen gestures, as time, fatigue, camera angle, muscle activation, and labeling efficiency may also vary.

The test blocks were equally divided into four groups in central order, with 74 pieces in each group. Table 8 retains the original time grouping results of the three baselines; the original EVFormer grouping values cannot be reconstructed after aggregation of the corrected 11.482°, and have been marked with "-" for review. It cannot currently be claimed that EVFormer outperforms vision in all four consecutive time groups.

The four groups have the same number of samples but cover different physical times: the first three groups are around 200–225 s, and the last group extends to 281 s. The labeling efficiency decreased significantly in the later period, resulting in the last 74 retained centers being more sparsely distributed. Therefore, equal sample grouping is suitable for describing the error of the model on the retained samples, but it is not suitable for representing the performance of all actions in each period of time. Table 8 lists the sample number and time range at the same time to avoid misinterpreting the sparse data in the later period as continuous and uniform observations.

## 6. Fine-Grained Experimental Results and Diagnostics

### 6.1 Detailed Preliminary Results

### 6.1.1 Per-Seed Results

Table 5 retains the seed-by-seed test MAE of the three baselines. 2026, 2027, and 2028 are random seeds for model initialization and data reading order, and do not correspond to different participants. The average of the three seed scores of the original EVFormer is inconsistent with 11.482°, and has been removed and marked with "-" for review. Reviewing requires matching checkpoints, test centers and frame-by-frame predictions, and cannot just add the same constant to each seed based on the mean change.

Table 5. Test MAE of each random seed (EVFormer sub-item to be reviewed)

| random seed | Vision MAE only | Myoelectric MAE only | Post Fusion MAE | EVFormer MAE |
|---|---|---|---|---|
| 2026 | 13.177° | 13.201° | 12.907° | — |
| 2027 | 13.233° | 13.484° | 13.575° | — |
| 2028 | 13.275° | 14.145° | 13.680° | — |

The test scores of the three seeds in the visual baseline are close, and the fluctuations in myoelectricity and late fusion are larger. EVFormer does not yet have seed-by-seed results corresponding to the revised principal values, and does not report its seed standard deviation, seed-by-seed improvement rate, or conclusion that all seeds won. Overall ranking and initialization stability should be verified separately, and three initializations cannot replace cross-user evaluation.

### 6.1.2 Descriptive Test Errors by Gesture Class

Table 6 groups the same batch of predictions according to the label.gesture_class field of the test center. The value is the group-by-group MAE average of the three random seeds in the original report, in degrees. The five categories cover a total of 296 test centers, and the overall MAE of EVFormer weighted by the number of samples is 11.482°, which serves as the basis for this unified main result. Category 22 has 142 samples, accounting for nearly half of the test set; category 27 has only 16 samples. Category labels are not used as model input, and the category table is used for post-training description. All samples are from the same participants and the same clip and cannot be considered as five independent estimates of population performance.

Table 6. Gesture category MAE and sample weighted overall results

| Gesture number (sample) | visual only | EMG only | late fusion | EVFormer |
|---|---|---|---|---|
| 21（51） | 12.834 | 12.715 | 12.889 | 12.989 |
| 22（142） | 14.392 | 14.657 | 14.333 | 12.076 |
| 23（59） | 12.197 | 13.064 | 12.862 | 10.235 |
| 24（28） | 13.158 | 13.637 | 13.262 | 11.041 |
| 27（16） | 8.084 | 9.142 | 8.740 | 6.783 |
| Sample Weighted Population (296) | 13.228 | 13.610 | 13.387 | 11.482 |

EVFormer achieved the lowest MAE among the four learning models on categories 22, 23, 24 and 27; category 22 was 12.076°, which was lower than vision 14.392°. Category 21 constitutes an exception: EVFormer is 12.989°, which is higher than visual 12.834°, electromyographic 12.715° and late fusion 12.889°, which are 0.155°, 0.274° and 0.100° higher respectively. Therefore, the overall best EVFormer does not mean the best in each category. Category 27 has only 16 samples, so it is not easy to infer that this category is universal. Figure 7 shows the class error and sample number side by side to preserve the local disadvantage of class 21 and the difference in class size.

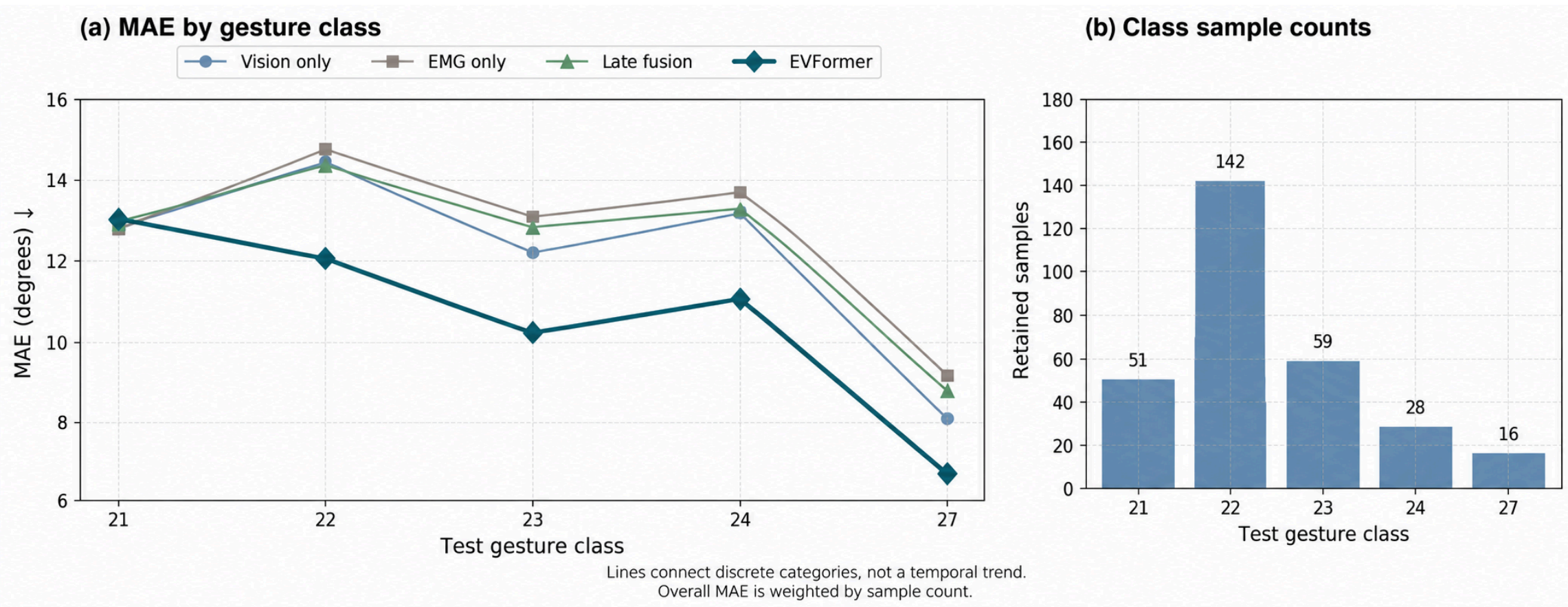

Figure 7. Error and sample composition by gesture class. Values are taken from Table 6. (a) Class-wise MAE for the four learned models. (b) Sample count for each class. Lines connect discrete classes and do not indicate a temporal trend.

### 6.1.3 Why Aggregation Matters

The number of samples in the five categories is 51, 142, 59, 28 and 16, totaling 296. Weighted by the number of samples, the overall MAE of EVFormer is (51×12.989+142×12.076+59×10.235+28×11.041+16×6.783)/296=11.4823378°, which is 11.482° to three decimal places. The visual, electromyographic and post-fusion calculations using the same method are approximately 13.228°, 13.610° and 13.387° respectively, which is consistent with the main table. Category 22 contributes the most to the overall; the macro average of the five categories is 10.6248°, which is not the main indicator of this article, and the macro average cannot be used to replace the sample weighted value.

## 6.2 Label Validity and Sample Population

### 6.2.1 Changes in Validity over Recording Time

Not all raw rows in Parquet have two-handed 44-dimensional regression labels. Table 7 counts the original number of rows in each time period according to the validity of labels generated by left and right hands and the validity of left and right wrist angles: nearly all of them are valid in the first 50 s, only about 3.3% in 250-300 s, and zero after 300 s.

This table does not include all model screening steps after sampling every 200 rows, input finiteness checking, and boundary removal, and cannot be directly compared with the final 1,964 model samples. The source file is about 345.7 s long, but the last remaining test center is at 281.0 s.

Table 7. Label validity statistics of original rows

| **Time interval (seconds)** | **The number of rows that satisfy the two-hand label and wrist angle rules** | **interval effective ratio** | **Impact on sample composition** |
|---|---|---|---|
| 0—50 | 99,983 | About 100.0% | Sufficient candidate centers |
| 50—100 | 83,988 | About 84.0% | Invalid interval occurs |
| 100—150 | 71,665 | About 71.7% | Available centers reduced |
| 150—200 | 79,603 | About 79.6% | Efficient recovery |
| 200—250 | 56,403 | About 56.4% | The test interval begins to become sparse |
| 250—300 | 3,266 | About 3.3% | There are only a few valid centers |
| 300—End of file | 0 | 0% | No valid two-handed target |

The above-mentioned effective rate is not the "success rate" of model prediction, but a statistic of whether the data label is available. Specific reasons for invalid labels may be related to hand motion capture visibility, the annotation generation process, or wrist angle calculation; existing fields alone cannot be attributed individually. Strict bimanual filtering will favor certain scenes that are more likely to receive both left and right hand labels. If such biases are related to the gesture or the degree of occlusion, the error calculated based on the held-out sample may also differ from the error during the complete motion process. Complete studies should report the proportion of left-handed, right-handed, and both-hands valid separately and examine their relationship to gesture family, timing, and image visibility. Figure 8 shows the effective proportion of tags in Table 7 along the time interval.

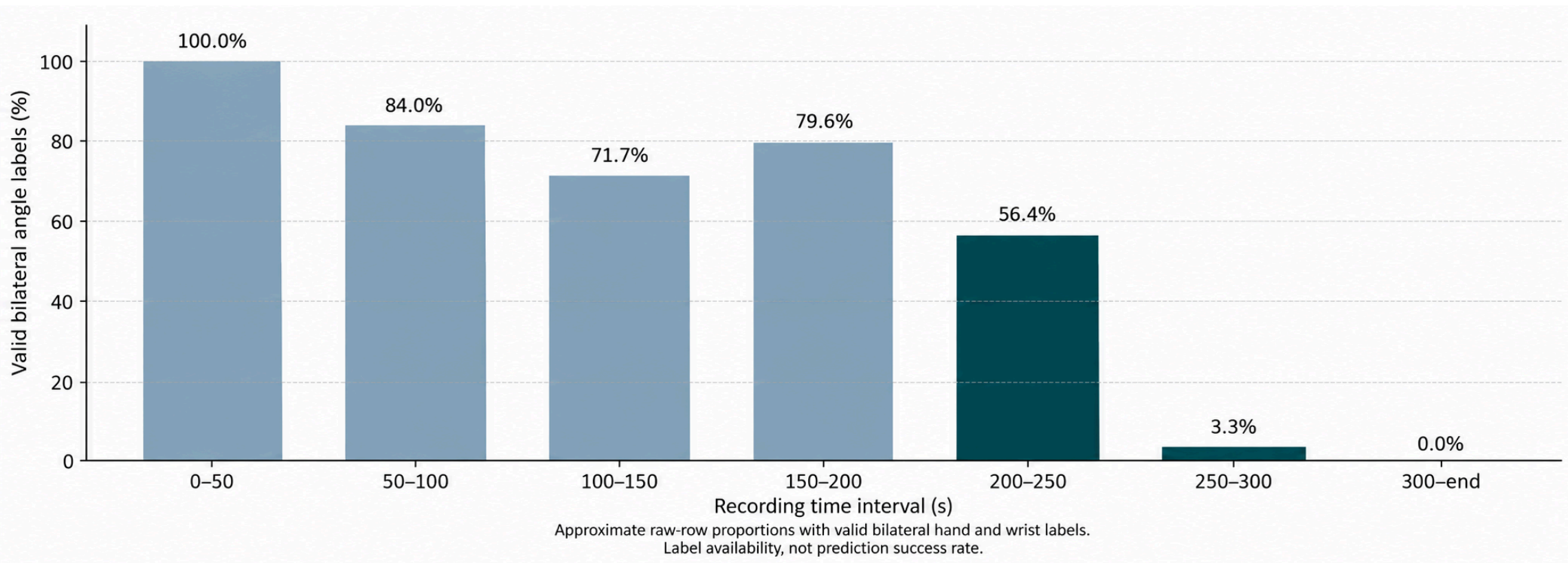

Figure 8. Proportion of valid bilateral hand and wrist labels over recording time. Statistics are based on the raw-row intervals in Table 7 and use the approximate percentages reported there. The figure shows label availability, not prediction success.

### 6.2.2 Independence and Statistical Units

A candidate center around 10 Hz does not mean that adjacent samples are statistically independent; the EMG windows of adjacent centers may also overlap. Removing samples from data block boundaries only ensures that the input windows between training, verification and testing do not overlap with each other, but cannot eliminate the correlation of the same collection session. The currently retained baseline seed standard deviation only reflects initialization changes; the seed standard deviation of EVFormer matching 11.482° has yet to be reviewed. This article does not report its confidence interval, nor does it express the degree of seed dispersion as a new user statistical accuracy.

On full EgoEMG, inferred units should be consistent with the research question. If the question is "Can a different user still work?", it should be aggregated by user or user-session, and user-level resampling or hierarchical statistics should be used to express uncertainty; if the question is "Change a type of gesture", the gesture group should be fixed first and avoid leakage of the same category during training and testing. Frame-level MAE is still a useful loss metric, but is not a substitute for a suitable independent unit of evaluation. While reporting the overall mean, per-user scatter or distribution can also be helpful in spotting critical failures for a small number of users.

## 6.3 Sample-Level Error Diagnostics

### 6.3.1 Definition and Unit of Pairwise Comparisons

Pairwise analysis requires calculating the 44-dimensional mean absolute error difference between EVFormer and the visual baseline for each seed and each test center. The original reported seed-level EVFormer mean value does not match the corrected main value. In the absence of the original prediction, the corresponding winning frame ratio cannot be confirmed. The relevant numerical values and distribution conclusions are not reported for the time being.

The review should retain the index of each center, the true target, and the predictions of both models, and use consistent seeds, labeling rules, and output units. The winning proportions were calculated within each seed before aggregating results across seeds; replications across seeds at the same test center are not independent human observations.

Overall MAE decreases can occur simultaneously with partial frame or category error increases. Table 6 currently shows partial counterexamples of category 21. Frame-by-frame revenue coverage also requires matching predictions and independent scene annotation verification, and cannot be derived from the overall ranking.

### 6.3.2 Equal-Count Temporal Groups in the Test Set

Table 8 divides the 296 test samples sorted by the center index into four consecutive groups of 74 in each group. The grouping indicators should first be averaged across the samples and 44 angles within the group, and then averaged across seeds; the equally weighted average of the four groups must reconstruct the overall test MAE. The current three baseline results are retained, and the EVFormer grouping value needs to be reviewed. Such diagnostics do not participate in model selection and cannot be attributed to specific events without action origin or visibility annotations.

Table 8. Continuous test grouping MAE (EVFormer sub-item to be reviewed)

| test group | Number of retained samples | Center time range (seconds) | Vision MAE only | Myoelectric MAE only | Post Fusion MAE | EVFormer MAE |
|---|---|---|---|---|---|---|
| Group 1 | 74 | 200.0—207.3 | 12.295 | 12.369 | 12.425 | — |
| Group 2 | 74 | 207.4—214.8 | 14.649 | 14.798 | 14.441 | — |
| Group 3 | 74 | 214.9—225.1 | 14.189 | 14.597 | 14.336 | — |
| Group 4 | 74 | 225.2—281.0 | 11.778 | 12.675 | 12.347 | — |

The number of samples in each group is the same, but the physical time span is different: the fourth group extends to 281.0 s, which is significantly longer than the first three groups. Equal sample grouping is suitable for describing retained samples and does not represent equal-length continuous observations. The EVFormer group data needs to be reviewed, and we currently do not report the improvement of each group or claim that all four groups have improved.

### 6.3.3 Raw Annotations Required for Further Diagnostics

The current file records images, electromyography, joint angles and coarse-grained gesture numbers, but there is no frame-by-frame artificial occlusion label that is sufficient to support "myoelectricity improved occlusion". If additional annotations are added to the picture in the future, it will at least be necessary to distinguish the visibility of the left hand, right hand and both hands, and also to record object contact, self-occlusion, boundary crossing and motion blur. Occlusion criteria should be developed before looking at model output; annotators should not decide whether a frame is "difficult" based on the error of a method. Only in this way can the preset occlusion subset results test the research motivation of EVFormer. The same applies to myoelectric quality: channel disconnection, electrode displacement and abnormal noise need to be defined independently of attitude error, and frames with large model errors cannot be directly called "bad signals".

In addition to the average error per group, a formal study should also show the distribution of pairwise differences by user and examples of failures. A single person can contribute a large number of adjacent frames, and if all the frames were blended directly, one long clip would overwhelm multiple short clips. For cross-user tasks, method differences can be calculated within each user first, and then user-level results are summarized. For cross-gesture tasks, method differences can be first summarized within predefined gesture families, and then inter-family stability can be compared. Section 6.3 is currently unable to complete statistics at these levels, due to the fact that the source image itself does not have a second user, rather than a lack of additional drawings.

## 6.4 Errors and Target Distributions by Hand and Joint Type

### 6.4.1 Left/Right Hand, Finger, and Wrist Errors

Table 3 uniformly lists the indicators of the left and right hands, fingers and wrist areas, and does not display the same data repeatedly. The regional values of the three baselines can be weighted by the output dimension to reconstruct their respective overall errors; the EVFormer regional result has been removed because 11.482° cannot be reconstructed. Currently only its overall and gesture category results are retained, and no left-hand and right-hand symmetry performance or specific joint contributions are inferred.

Regional and joint-by-joint analyzes need to be recalculated from the matched original predictions. The weighted summary of the 40 dimensions of the fingers and the 4 dimensions of the wrist should be consistent with the main index; the regional mean cannot explain the direction and amplitude of a single degree of freedom, nor can it be interpreted as a selective contribution to a specific muscle group.

### 6.4.2 Descriptive Statistics of Training, Validation, and Test Targets

In order to explain why the training mean attitude reference has reached 13.320°, Table 9 lists two descriptive quantities of the target angle in each time block: "Average absolute angle" is the average |angle| on the sample and 44 dimensions; "Average dimension-wise standard deviation" first calculates the overall standard deviation for each angle within the block, and then averages across dimensions. They reflect the scale of the target distribution, not the error of the prediction model, nor do they measure label accuracy. The wrist only occupies 4 dimensions, so finger and wrist standard deviations are listed separately to prevent the overall mean from masking regional differences.

Table 9. Descriptive statistics for training, validation and testing objectives

| data block | Number of samples | mean absolute angle | mean dimension-wise standard deviation | Finger-by-dimension standard deviation | Wrist dimension-wise standard deviation |
|---|---|---|---|---|---|
| training | 1,375 | 20.154° | 15.524° | 15.171° | 19.057° |
| Verify | 293 | 19.874° | 11.142° | 10.273° | 19.838° |
| test | 296 | 21.350° | 12.731° | 12.099° | 19.048° |

These within-block statistics do not directly give the mean difference between blocks, so the increase in MAE from validation to testing cannot be explained by Table 9 alone. The training block covers a variety of older gestures, and the dimension-wise standard deviation is higher than the validation block; but the standard deviation also depends on the proportion of each gesture in the block. To illustrate the specific failure of the model in the face of unseen gestures, it also needs to be reviewed on the gesture-level and user-level distributions of the complete data. For this excerpt, the training mean reference is close to the test MAE of the visual model, suggesting that this simple reference must be maintained to avoid over-interpretation of the numerical advantages of complex models in a narrow distribution.

## 6.5 Model Capacity, Fairness of Comparisons, and Computational Scope

### 6.5.1 Instantiated and Trainable Parameters

The code instantiates the same EVFormer class in all modes, so each object contains 171,116 learnable parameters; but the forward branch determines which modules actually participate in output and backpropagation. We use a constructed input to perform the forward and backward calculations of each mode, and count the number of parameters with non-empty gradients. The results are shown in Table 10. This count only indicates which parameters are accessed by the computational graph in the current implementation and does not represent inference FLOPs, memory bandwidth, or real training time; for example, all modes perform the forward of the two-pass encoder, even if the single-modal output does not use the other branch.

Table 10. Total parameters of the model object and parameters participating in the gradient

| training mode | Model object total parameters | parameters with gradients | Difference from vision-only gradient parameter |
|---|---|---|---|
| visual only | 171,116 | 62,572 | 0 |
| EMG only | 171,116 | 69,996 | +7,424 |
| late fusion | 171,116 | 137,836 | +75,264 |
| EVFormer | 171,116 | 171,116 | +108,544 |

There is a difference of 33,280 parameters actually participating in the gradient between late fusion and EVFormer. Therefore, the accuracy differences shown in Table 2 compare the two configurations in the current implementation, but are not yet strictly equal-volume ablation, and the entire gain cannot be attributed to cross-attention alone. A complete study should add visual and post-fusion controls for parameter matching and keep training data, optimizers, image cropping, and input windows consistent.

### 6.5.2 Theoretical Interaction Size and Measured Latency

The current visual sequence has 16 tokens, and the electromyographic sequence has 25 tokens. The four-head cross attention forms 16×25×4=1,600 query-key scores in each direction, and the total number of sequential two-way is 3,200. This amount does not include two-pass encoding, linear mapping, regression, or video decoding costs. 200 ms represents the length of the EMG history used in regression. The streaming rolling cache can continuously provide this history. It cannot be directly regarded as the 200 ms waiting or end-to-end delay added for each prediction. Formal systems require measurement of synchronization, preprocessing, inference, and output transmission times on explicit hardware.

Deployment scenarios also need to consider caching strategies. If each new center reprocesses 400 raw EMG points, the calculation and data handling will be different from maintaining a rolling cache online; if the image is 60 fps and the prediction center is about 10 Hz, there may still be queuing and decoding overhead in the same image acquisition link. If EMG filtering uses future sample points, you cannot simply define the input as "200 ms before center" and claim causality. This article does not conduct online delay measurements based on real devices, so the above content is a requirement for the measurement protocol, not a performance result.

### 6.5.3 Minimum Reporting Unit for a Complete Evaluation

A complete EgoEMG study would need to report the number of users, number of gestures, number of test samples, and overall, left- and right-hand and wrist MAE for each model on at least three official divisions. For model comparison, the pairwise differences of EVFormer and baseline can be calculated within each test user, and then the user-level distribution can be displayed; if a certain partition has unseen gestures as the main generalization axis, the results by gesture family can be added. Success criteria should be stated before looking at the final test set, such as requiring consistent and meaningful improvements over a matching fused baseline across multiple partitions, rather than picking out one of the favorable numbers after the fact. Negative results should also be published together with training configurations, running seeds, and prediction files to prevent new studies from repeating unverifiable conclusions.

## 6.6 Audit Plan and Scope of Controlled Input Perturbations

### 6.6.1 Fixed Input Transformations

This section explains the fixed input transformation and review requirements listed in Table 4 and does not consider them to be experimental results that have been matched to modified principal values. The review should maintain the original 1,375 training, 293 validation and 296 test centers, use clean validation data to select checkpoints, and then use the same checkpoint to evaluate all applicable conditions. Test perturbations are not used for early stopping or parameter adjustment, and no new participants, gestures or angle labels are added; relative changes can only be re-summarized after clean recording and perturbation prediction according to the seed.

RGB crop is scaled to 64×64, with values falling in [0,1]. The central occlusion zeroes the 32×32 area with height and width coordinates of 16-47, covering a quarter of the area. The rectangle is fixed in position and does not depend on the outline of the hand. It may cover the fingers or mainly cover the background. The 5×5 mean blur uses stride 1 and zero padding 2, which is not exactly the same as reflection padding. Zeroing the entire image will set all pixels to zero; whether it belongs to a missing modal condition not seen in training should be judged based on the actual implementation of modularity dropout.

EMG inputs were first normalized channel-by-channel using the mean and standard deviation of the training block and clipped to [−8,8]. Zeroing one channel on each wrist fixedly selects the 0th and 8th channels, i.e. one channel for each wristband; since zeroing occurs after normalization, "zero" corresponds to an input near the training mean rather than the physical sensor output of 0 V. Gaussian noise uses a fixed random source 1337, adds independent noise with a standard deviation of 0.5 to each normalized value, and then clips back to [−8,8]. Setting the entire EMG circuit to zero will set all 100×16 standardized values to zero, which means replacing the input with the training center value and does not represent the waveform of the real disconnected electrode. The combined conditions apply central occlusion and two-channel zeroing at the same time, or apply the entire channel image and the entire channel EMG zeroing at the same time. Figure 9 shows the above transformation as a schematic input, without involving the perturbation problem that has yet to be reviewed.

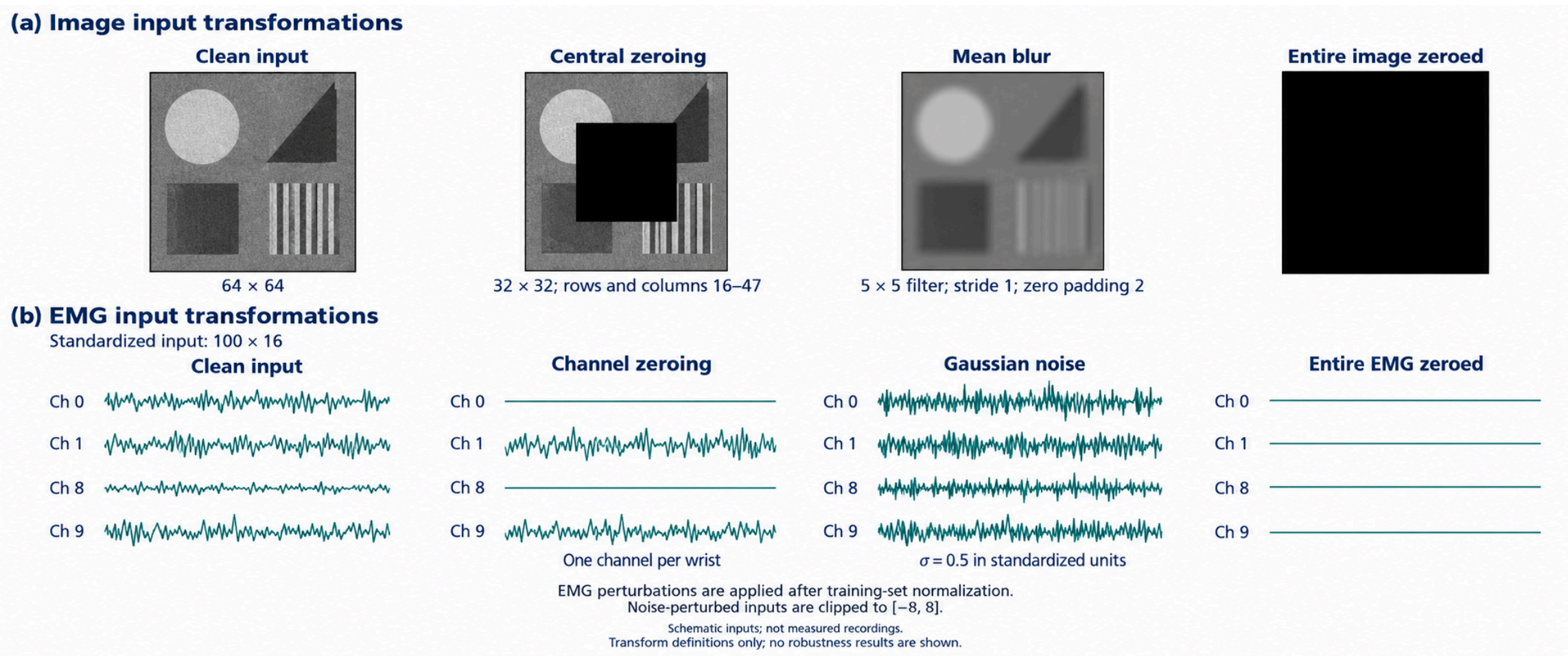

Figure 9. Controlled input perturbations (schematic). (a) Central image zeroing, mean blur, and whole-image zeroing. (b) EMG-channel zeroing, Gaussian noise, and whole-EMG zeroing. The textures and waveforms are illustrative rather than measured data, and no robustness results are shown.

### 6.6.2 Checkpoint Consistency and Per-Seed Audit

Table 5 uniformly displays the seed-by-seed test results, without repeatedly listing the original EVFormer seed value or the unreviewed winning frame ratio. During review, the clean, perturbed and paired results of each seed must come from the same checkpoint and the same test center, and the checkpoint identification, input transformation parameters and random source are recorded.

Each disturbance needs to retain its sub-scores one by one and calculate the difference with the corresponding clean scores. The corrected main values cannot be mixed with the absolute values of old disturbances from unknown sources. These details are not currently archived, and perturbation means, relative changes, or inter-seed rankings are not reported in this article.

### 6.6.3 Checkpoints, Result Files, and Interpretive Limits

Existing archived material does not contain checkpoints, per-by-seed JSON, raw frame-by-frame predictions, or run logs matching 11.482°. Table 6 provides a summary basis for overall revisions but does not replace the original forecast. After supplementary filing, the homology of the overall, regional, category, time grouping and disturbance results needs to be reviewed; "—" in this manuscript indicates that the corresponding EVFormer sub-items have yet to be reviewed.

The training settings include a modularity dropout of 0.10, so it cannot be claimed without verification that all test zero conditions never occur in training. The review should indicate which modalities, alternative values, and trigger probabilities were discarded by the training, distinguishing between missing modal conditions covered by training and unseen local noise or input changes. Software zero setting does not equal real occlusion, electrode shift or clock mismatch; if we discuss anti-occlusion or anti-fault performance, we also need to compare it with real conditions and a baseline that accepts the same perturbation.

## 7. Discussion and Future Evaluation

### 7.1 Main Findings

Under the conditions of single-segment preprocessing and division in this article, the overall test MAE of EVFormer is 11.482°, which is lower than the visual 13.228°, electromyography 13.610°, post-fusion 13.387° and training mean posture 13.320°. It is the best overall performance model in the current comparison. The relative vision was reduced by 1.746° and improved by 13.20%; the relative late fusion was reduced by 1.905° and improved by 14.23%. The gesture category results include local disadvantages of Category 21, and the seeds, regions, time groupings and perturbation data to be reviewed are not used as additional evidence of advantages.

The current experiments show a reduction in error, but existing controls cannot single out the mechanism by which the gains arise. Fixed 64×64 cropping, short EMG history, single-participant data size, randomly initialized visual backbone, cross-modal interactions, and dynamic gating may all affect the results; these are study factors for subsequent controlled ablation, not causal explanations that have been confirmed in this experiment.

### 7.2 Scope of Evaluation Relative to Baselines and Dataset

The current results support the overall accuracy advantage of EVFormer in this article's internal comparison, and it cannot be claimed that it is the best in each category or has outperformed all published methods. The official multi-user division of EgoEMG has not yet been adopted, and the matching residual fusion baseline has not yet been completed. The relevant comparison still needs to be verified after unifying the data, training budget and capacity.

### 7.3 Limits of Interpretation

This article has only one participant and one recording, and no independent test users. Time blocking ensures that the input windows across data blocks do not overlap, but it cannot eliminate the correlation of the same session; category distribution, fixed clipping, third-party mirror versions and potential offline electromyography processing all limit extrapolation. EVFormer sub-items and perturbation results have yet to be reviewed against the original predictions, and currently cannot support initialization stability, true occlusion robustness or sensor failure conclusions. Source file hashes can only identify data identity and cannot replace experimental logs, predictive archiving, and independent recurrence.

### 7.4 Sequence of Follow-Up Experiments

If the next round of research allows only a small number of controlled experiments, it should first be determined which hand areas are actually seen by the current weak vision baseline. Although a fixed crop is easy to reproduce, the pixels occupied by the finger may be very few; the background in the same crop may also create shortcuts with the recorded scene. You can compare fixed crops, hand boxes determined by the training set, and high-resolution crops while holding the rest of the conditions constant, and record detection failures and dropped frames one by one. If the visual model is still close to the training mean pose baseline, it is difficult to explain the benefits of increasing the fusion complexity.

The second step should be to review synchronization and input disturbances. Use checkpoints consistent with the 11.482° main result to re-evaluate fixed image occlusion, blur, channel zeroing, EMG noise, and full channel zeroing, and record the relationship between training modality dropout and test conditions. Clean and disturbed predictions of the same test center are saved seed by seed, and the same applicable conditions are applied to the baseline; the true electrode displacement and multi-user signal quality still need to be collected separately.

The third step is to adjust the cross-modal structure. Sequential cross-attention introduces more effective training parameters than late fusion, and also introduces learnable assumptions about the relationship between time slices and image regions. If improvements come from more parameters than from information interactions, structural claims are not supported. Formal ablation should control the number of effective parameters and training budget, and set up single-modality or parallel fusion models of equal scale; at the same time, compare cross-attention order, retain only one direction, and different myoelectric history lengths. Any hyperparameters selected through the validation set must be evaluated on a predetermined test split and cannot be screened repeatedly on the same test block.

## 7.5 Prespecified Evaluation on the Full Dataset

### 7.5.1 Data Splits and Comparisons

Before viewing the test results of complete data, you should first fix and archive EgoEMG's official three sets of "gesture", "user" and "both at the same time" division files. Leave participants or gestures on the axis unavailable for training normalization, checkpoint selection, image cropping adjustments, or data augmentation selections. For the same sample, all candidate models should receive the same center frame image and synchronized EMG history. Controls include at least: a vision-only model using the same visual backbone as EVFormer, an EMG-only model with the same EMG encoder, late fusion, EVFormer sequentially removing one cross-attention direction, and a residual fusion baseline of EgoEMG [1]. WiLoR [10], a stronger hand-specific vision model, can also be added, but its pre-training source and capacity must be stated.

Primary results should be based on the officially defined test set joint angle MAE in degrees and summarized in the original benchmark manner. Before calculating the cross-division average, the scores for "Gesture", "User" and "Both" are reported separately. Each set of partitions also describes the participants, gestures, and number of valid gesture frames, and provides per-user mean and dispersion, left and right hands, fingers and wrists, and results for a number of prespecified gesture families. If you don't see a significant increase in user error, you can't claim method improvement based on just an overall average.

### 7.5.2 Robustness and Ablation Studies

Occlusion experiments should use image quality variables that are independent of prediction error, such as manually verified visibility annotations or preset occlusions after model freeze. All available visual methods should accept the same perturbation. EMG channel discarding, noise and timestamp jitter require clear amplitudes. Modality dropout is currently used for training, but the missing modal results accompanying the modified principal values have yet to be reviewed; it cannot be claimed that the missing modal capability has been verified based on the gating structure or training configuration alone.

Attention blocks should be removed or replaced while controlling the number of effective parameters as much as possible. Valid comparisons include: vector-level post-fusion, visual query EMG only, EMG query vision only, two-way interaction, and residual pose correction head. The EMG history length and image cropping resolution should only be changed one factor at a time. These experiments can distinguish whether gains come from true cross-modal interaction, greater model capacity, or simply from more appropriate image cropping or longer EMG context. The results of these full data experiments are not yet available for this article.

### 7.5.3 Reporting Requirements and Failure Criteria

The evaluation should retain all prespecified random seeds and their distributions, rather than picking the favorable ones just once. Confidence intervals should be calculated based on the participant or acquisition group and not on adjacent video frames as independent observations. Training time, video memory, number of parameters, and inference latency should be measured under explicit hardware conditions. If EVFormer's error on a prespecified subset of occlusions or unseen users is no worse than the matched visual and residual fusion control, then the hypothesis that EVFormer is helpful should be rejected for this configuration. The negative results on the complete data still shed light on the conditions under which bilateral EMG fails to provide additional value for strong first-person vision.

## 7.6 Implementation Sequence for a Complete Study

### 7.6.1 Data Verification and Baseline Reproduction

The first step is to obtain the official complete data of EgoEMG, split files and codes consistent with its version, and record the version number and source file check value. The loader should sample the video center frame, left and right EMG windows, the order of the 22 left and right hand corners, radians to degrees conversion, and validity masks. Random checks cannot be done only on the training set: each official partition should confirm that the user and gesture

IDs are consistent with their stated retention conditions. If you use off-the-shelf detectors or annotations for image cropping, checkpoints and failure handling strategies must also be recorded to avoid manual adjustments on the test set from entering the model selection process.

The second step should be to reproduce at least one visual, one EMG and one residual fusion baseline according to the configuration published by the original author, and check that the output indicators are within a reasonable range with the original paper. If the recurrence values are obviously different, first locate the target unit, crop size, frame filtering, sample weighting and pre-training source instead of adjusting the EVFormer hyperparameters immediately. After completing the baseline reproduction, run EVFormer and post-fusion ablation with the same training budget. The comparison obtained in this way can answer "whether cross-attention brings gains" rather than "whether the two training pipelines are different".

### 7.6.2 Prespecified Outcomes and Resource Reporting

The main MAE, the three partitions, the models listed in the main table, the number of random seeds, and the definition of occlusion and EMG noise subsets should be fixed before the formal experiment begins. The main table should list the number of samples, the number of users, the mean and the degree of dispersion one by one; the appendix should list the left and right hands, fingers, wrists and each degree of freedom. If exploring multiple history lengths, image resolutions, or encoder sizes, you should separate the exploration set from the final test set and fully disclose the settings you tried. The amount of training computation, peak video memory, effective model parameters, and end-to-end latency are also part of the value of the method, especially in the context of wearable devices.

Synchronization errors should also be stress tested. For example, applying predetermined offsets of ±20, ±40, and ±80 ms to the EMG input without changing the true tag timestamps; rapid degradation in model performance may indicate the need for tighter clock synchronization during deployment. The image side can use a fixed area of occlusion or blur to detect visual degradation; the EMG side can use a fixed probability of channel discarding and noise within a certain signal-to-noise ratio range to detect sensor degradation. All methods should accept the same degree of perturbation, and the intensity of the perturbation cannot be temporarily modified due to poor performance of a certain model.

### 7.6.3 Three Possible Conclusions from a Complete Study

If EVFormer is stably lower than the matching visual, post-fusion and residual fusion baselines in all three official divisions, and there are benefits among unseen users, it can be discussed as being more effective in this setting. If it is only beneficial on a subset of occlusions but not on normal images, the conclusion should be restricted to predefined occlusion conditions and the overall cost reported. If it does not improve overall or in key subsets, the conclusion should be that the current cross-attention design is not supported by evidence; further analysis can be done to determine whether the image backbone is too weak, the EMG difference across people is too large, or the two-way attention is overfitting. Regardless of the outcome, the original forecast, split ID, and full configuration should be retained so that others can review the summary table rather than just seeing the culled final numbers.

## 7.7 Conceptual Boundary between Pose Estimation and Movement-Intent Prediction

### 7.7.1 What the Current Model Predicts

The goal of this paper is the joint angles of the hands at the corresponding moment of the current image, not discrete gesture labels, nor future actions before the movement begins. Although the EMG input is taken from 200 ms before the center moment of the image, the image itself is from the current moment; the target angle is also aligned with the current moment. Therefore, the model uses "past EMG + current image" to estimate the current hand posture. This setting alone cannot claim to predict movement intentions in advance, nor can it directly convert the average angle error into milliseconds for the control device to respond in advance. Even if myoelectric activity precedes visible movement in some actions, whether there is a stable amount of lead available will require dedicated temporal alignment experiments.

If future research targets motor intention, the operational definition of intention should first be stated. For example, one can define joint angle trajectories Δ milliseconds into the future, future action categories, or “predict a specific motion direction before the action starting point.” Each definition requires different labels, origin detection rules, and error metrics. Future trajectory prediction must strictly limit the input to only come from the prediction moment and before, including video frames, EMG filtering and any normalized state; otherwise the so-called "prediction in advance" may be future information leakage. The starting point of the action cannot be selected after the fact based only on the model score, but should be determined by an independent annotation or pre-registered algorithm.

### 7.7.2 Evaluating Practical Value for Control

The joint angle MAE is just an offline reconstruction accuracy indicator. For prosthetics, rehabilitation or virtual interaction devices, the actual value also depends on whether the continuous output jitters, the user's ability to adapt in a closed loop, system end-to-end latency, wearing comfort and stability of repeated electrode placement. A model with lower MAE offline may not necessarily be more suitable for online use if it requires high-latency video decoding or frequent manual calibration. On the contrary, a lightweight model may be more practical due to its stable response even if the angular error is slightly higher. The above issues are beyond the scope of this preliminary experiment, and the offline results of a single fragment cannot be used to replace user research or equipment testing.

Therefore, EVFormer should currently be called a visual-myoelectric hand pose estimation model. If it is developed into a motion intention recognition system, the prediction target, action starting point and verification protocol must be separately defined. The MAE of the current continuous angle regression cannot be directly used as discrete gesture recognition or future action prediction performance.

## 8. Conclusion

This paper introduced EVFormer, which exchanges information between visual spatial tokens and bilateral-wrist EMG temporal tokens through residually scaled sequential bidirectional cross-attention, then uses feature-wise gating to regress 44 bimanual joint angles. In a unified internal comparison based on one participant and one recording, EVFormer achieved a sample-weighted test MAE of 11.482 degrees, lower than vision-only (13.228 degrees), EMG-only (13.610 degrees), late fusion (13.387 degrees), and the training-mean pose (13.320 degrees). This corresponds to relative improvements of 13.20% over vision-only and 14.23% over late fusion. EVFormer performed best in four of the five test gesture classes, while retaining a local disadvantage for class 21. Per-seed, regional, temporal-group, and perturbation results that do not match the corrected aggregate still require verification against the raw predictions and are not used to support the present conclusions. The available results support an overall accuracy gain under this experimental setting; generalization across users, robustness to real input degradation, and online performance require further study.

# Appendix A: Data Fields and Angle Semantics

## A.1 Input and Target Fields

Parquet clips save EMG, motion capture related information, generated pose annotations, and their association with video frames. The pre-experiment reads observation.emg.left_filtered and observation.emg.right_filtered, each line has 8 values; read observation.images.webcam.frame_index to select the current video frame. The generated finger angle arrays are generated.joint_angles.left and .right, with 20 corners for each hand. The two-element field generated.label_valid indicates whether the labels generated for the left and right hands are valid. Another four fields give the pitch and yaw of the left and right wrists, and require the corresponding angles_valid flag to be true. label.gesture_class is only used for division diagnosis after training is completed. It is neither a model input nor involved in checkpoint selection.

The output sequence follows EgoEMG's 22 degrees of freedom convention [1]: indexes 0-3 belong to the thumb, 4-7 to the index finger, 8-11 to the middle finger, 12-15 to the ring finger, 16-19 to the little finger, and 20-21 to the wrist. Each group of four finger angles corresponds to the movement of metacarpophalangeal and interphalangeal joints in the release representation; index 20 is wrist flexion and extension, and 21 is radioulnar deviation. The left and right hands use the same indexing semantics, and the right hand is shifted back 22 bits overall in the 44-dimensional total vector. Table A1 summarizes the slices used by the indicator.

Table A1. Hand output index and indicator dimensions

| indicator area | left hand output index | Right hand output index | Total dimensions |
|---|---|---|---|
| Finger corners | 0—19 | 22—41 | 40 |
| wrist angle | 20—21 | 42—43 | 4 |
| complete hands | 0—21 | 22—43 | 44 |

## A.2 Unit Conversion and Validity Checks

Finger angles in the source data are measured in radians, and wrist pitch and yaw fields are expressed in degrees. If mixed directly, both the training target and the final MAE will be seriously distorted. Preprocessing therefore only performs radians to degrees conversions for the 20 finger corners of each hand, and the spliced wrist values remain in their original units. One unit test checks this step using a preset 90° finger angle and a specified wrist angle; another test confirms that the EMG window consists of the sampling point just before the center row. These values are only used for checking the program and do not enter the paper's training or test samples. Both errors can result in apparently reasonable but invalid scores, and therefore need to be explicitly guarded against.

The pre-experiment requires that both left and right hand labels and left and right wrist angles are valid. This strict ambidextrous rule reduces the available sample but ensures symmetrical validity conditions for the left and right hands at the time of assessment. Future full-data studies could use hand-by-hand validity masking to retain information from samples with only single-hand labels; it would then be necessary to account for how masking changes the sample population. The current joint prediction of two hands means that even if one of the hands is available, as long as the label of the other hand is invalid, the entire observation is still deleted.

# Appendix B: Data Splits and Gesture Composition

## B.1 Exact Sample-Selection Procedure

The nominal sample rate of the source stream is 2 kHz. The candidate center starts at row 400 and advances 200 rows at a time; the EMG history of length 400 ends just before the center. Candidate centers need to pass four label validity checks and camera frame index checks. The remaining centers are strictly incremental. We set cut points at 70% and 85% of the sequence, and then removed samples close to the boundary that would cause the 400-line EMG history to overlap with adjacent blocks. All models used the final 1,375/293/296 split; 4 otherwise valid samples were removed due to boundary spacing. This partitioning is deterministic and is not affected by the training random seed.

The actual length of the three data blocks differs from the 70%/15%/15% sample ratio because the label efficiency varies along the segment. The source file extends to approximately 345.7 s, with the final retention test centered at 281.0 s. The test cannot be interpreted as covering the entire acquisition process. Formal archiving should save each central row index, video frame index and corresponding prediction to check the homology of all grouped results.

### B.2 Gesture IDs in Each Data Block

Table B1, calculated from the gesture category field at the center of the valid sample, illustrates why this test score cannot simply be interpreted as "extrapolation to future time." The labeled gestures for training and testing barely overlap. -1 indicates that no gesture label is assigned, and it only appears in the training block in this filtered sample. The counts in the table are neither statistics for all rows of the original clips, nor statistics for EgoEMG's official test divisions.

Table B1. Data division, gesture categories and total number of samples

| data block | Gesture number marked | Total number of valid samples | The number of gesture samples is not marked |
|---|---|---|---|
| training | 8—19 | 1,375 | 47 |
| Verify | 19、20、21 | 293 | 0 |
| test | 21、22、23、24、27 | 296 | 0 |

The total number of valid samples in Table B1 includes gesture unlabeled samples. The 47 unlabeled samples in the training block are a subset of 1,375 samples, and the remaining 1,328 have gesture category labels. Both verification and testing have category labels; the counts of the five categories of testing are 51, 142, 59, 28, and 16 as shown in Table 6. Full class statistics should be archived with the original forecasts, currently not referencing local results files that are not yet available. The category ratio of short clips will affect the overall MAE, and strict generalization of gestures still requires official multi-participant classification [1].

## Appendix C: Implementation Details and Audit Trail

### C.1 Lightweight Model Configuration

The pre-experimental network uses 64-dimensional tokens, four attention heads per layer, 16 visual tokens and 25 electromyographic tokens. The image backbone uses 5×5 and 3×3 convolutions in sequence; the EMG backbone uses one-dimensional convolutions with temporal kernel lengths of 7 and 5 respectively. Each modality has a layer of Transformer encoder with a feedforward sublayer of 128 dimensions and a dropout of 0.1. The bidirectional cross-attention residual scaling factor is 0.1; the gate is a 128→64→64 perceptron, ending in sigmoid, and the last layer bias is initialized to +2.0. The final regression head consists of layer normalization, a 64→64 GELU hidden layer, and a 64→44 output layer. The fully constructed module has a total of 171,116 parameters; the parameters actually involved in calculations vary between different training modes.

### C.2 Training and Evaluation Implementation

The TensorDataset in the code saves the decoded image cropping, normalized EMG sequence, and target expressed in degrees. Image pixels are scaled to 0-1. The mean and standard deviation of each channel of EMG are only estimated from the samples and time positions of the training block, and the test statistics are not involved. The training loss is applied to the target divided by 90, and the prediction is multiplied by 90 before reporting the angular MAE. Each mode and seed is reloaded to verify the optimal checkpoint before calculating test metrics. Early stopping relies only on the validation MAE, so test scores do not participate in model selection.

This article uses the sample weighted value of 11.482° in Table 6 to unify the overall results of EVFormer. The original seeds, regions, time groupings and disturbance values that are inconsistent with this main value have been removed; the matching server-by-server seed JSON, checkpoints, original prediction arrays and logs have yet to be archived. It is necessary to supplement the server hardware, software version and operating configuration at the same time, and verify that the overall score can be reconstructed after weighting each group. Products that are not currently archived are not used as evidence that they can be directly recalculated locally.